\def\year{2022}\relax
\documentclass[letterpaper]{article} 
\usepackage{aaai22}  
\usepackage{times}  
\usepackage{helvet}  
\usepackage{courier}  
\usepackage[hyphens]{url}  
\usepackage{graphicx} 
\usepackage{natbib}  
\usepackage{caption} 
\DeclareCaptionStyle{ruled}{labelfont=normalfont,labelsep=colon,strut=off} 
\usepackage{algorithm}
\usepackage{algorithmic}

\usepackage{newfloat}
\usepackage{listings}
\floatstyle{ruled}
\newfloat{listing}{tb}{lst}{}
\floatname{listing}{Listing}
\title{Model-Free RL Agents Demonstrate System 1-Like Intentionality}
\author {
    Hal Ashton,\textsuperscript{\rm 1}
    Matija Franklin, \textsuperscript{\rm 2}
}
\affiliations {
    \textsuperscript{\rm 1} Computer Science, University College London\\
    \textsuperscript{\rm 2} Experimental Psychology, University College London\\
    matija.franklin@ucl.ac.uk
}

\usepackage{bibentry}

\begin{document}

\maketitle

\begin{abstract}
This  paper argues that model-free reinforcement learning (RL) agents, while lacking explicit planning mechanisms, exhibit behaviours that can be analogised to System 1 ("thinking fast") processes in human cognition. Unlike model-based RL agents, which operate akin to System 2 ("thinking slow") reasoning by leveraging internal representations for planning, model-free agents react to environmental stimuli without anticipatory modelling. We propose a novel framework linking the dichotomy of System 1 and System 2 to the distinction between model-free and model-based RL. This framing challenges the prevailing assumption that intentionality and purposeful behaviour require planning, suggesting instead that intentionality can manifest in the structured, reactive behaviours of model-free agents. By drawing on interdisciplinary insights from cognitive psychology, legal theory, and experimental jurisprudence, we explore the implications of this perspective for attributing responsibility and ensuring AI safety. These insights advocate for a broader, contextually informed interpretation of intentionality in RL systems, with implications for their ethical deployment and regulation.
\end{abstract}

\section{Introduction}
Identifying intent in an AI actor can aid with explainability as well as determining the legality of its actions. Evidence also suggests that it will have a bearing on lay-judgements of responsibility and culpability in the event that its actions cause some harm. Certain harms such as deception are defined in terms of intent, which means that an AI system or its designers prevent that harm, by identifying intent in the first place.

This article presents the view that intent can exist in model-free Reinforcement Learning (RL) agents. This is despite the fact that, on one view, they operate in a true system 1 way. This position is perhaps unintuitive given our a folk understanding of intent which suggests some sort of planning effort, and a causal model of the world. The question of whether intent is a system 1 or system 2 process has not been widely considered.. One reason perhaps is that in psychological research, the dichotomy is not as widely accepted as its general popularity would suggest for reasons we will discuss in this article. Another reason is perhaps that questions of intent determination, when stakes are high at least, have almost exclusively concerned humans (or organisations composed of groups of humans). Pinning intent down to a system 1 or type 2 mental process hasn’t really mattered because we agree that most humans are capable of intent and habitually think in both ways. AI systems and in particular RL agents force us to ask questions about which cognitive processes allow intent because they do not ‘think’ like humans and yet might be given real responsibility to act on their behalf. Whereas with human decision making one can make an argument about multiple decision processes informing any action, this is not the case with typical RL agents which just react to a situation by looking up an action according to their policy function. 

The article will proceed as follows; firstly we will briefly introduce model and model-free reinforcement learning. Next we will consider the emergence of the dual process account, discuss its criticisms and then consider intent within the context of system 1 thinking. We explore why intent is of interest when defining acceptable behaviour and look at how the law considers intent. Finally we argue that model-free should not meet intent free but discuss why an explanation of intent does need a model. Here we reference recent advances in safe-RL that use a structure known as a shield and draw a parallel with System 2 thinking.

\section{Model and Model Free Reinforcement Learning}
Reinforcement Learning (RL) agents have in recent years demonstrated extraordinary success in a number of many different applications which require sequential decision making subject to changing conditions. 
Whilst there are many different types of RL algorithms in existence but, they generally have a few things in common. They perceive the world as a vector of measurable states and they have available to them at any time a choice of actions which alter the world in some way. An exogenous reward function gives them feedback for the actions and the states that occur. The task of a RL agent is to maximise this stream of rewards by learning a policy which is a function that maps the state of the world to an action or a distribution of actions. The stream of rewards is the signal is through which the agent is able to learn how to improve its policy (behaviour). 
Some RL agents learn \cite{hafner_learning_2019} or are imbued with a perfect model of the world (\cite{silver_mastering_2017}) with which to understand the outcome of their actions and plan accordingly. Often this planning process is performed with some variety of Monte Carlo Tree Search (MCTS).
In contrast, Model-Free (MF) RL agents do not learn or have a model of the world. Essentially this class of agent learns through trial and error. Surprisingly this class of RL algorithm has proven itself to be remarkably good at many tasks \cite{shen_towards_2022}. For the practitioner they are attractive because there no modelling effort is required. However, their success is predicated on the ability to gain experience at low cost. In truth, methods which combine both learning methods have existed for a long time \cite{sutton_integrated_1990} and the barrier between the two is blurring \cite{ha_world_2018,collins_beyond_2020}. 
At the point where a RL agent has finished learning, its policy function is made static. The agent then chooses whatever action (or draws from a distribution of actions) that its policy function decides is appropriate given the current state of the world. The process of choosing an action is therefore one of looking up; no reflection on the part of agent takes places. It seems appropriate to equate this as system 1 thinking or (thinking fast). 

\section{The emergence of the dual process account}
The motivations of the dual-process account of human decision-making can be traced back to Bernoulli’s attempts in the 18th century to understand decision-making under risk \cite{mishra2014decision}, the study of utility and how this varied between people. This informed expected utility theory (EUT) of \cite{von1994theory} which formalised the idea of people’s preferences over choice. They showed that adherence to five axioms was sufficient to prove the existence of a utility function for any set of preferences. Adherence to these axioms was then interpreted as a way of being ‘rational’ \cite{suhonen2007normative}. Inconveniently, when it was found that people’s subjective judgements of probability differed from the real ones, \citet{savage1954foundations} created Subjective Expected Utility theory (SEUT). Even under these weakened axioms, \citet{kahneman1979prospect} found that under laboratory conditions, people rarely adhered to them, making them ‘irrational’.

\citet{simon1957behavioral, simon1979rational} developed a theory of ‘bounded rationality’ whereby they exhibited satisficing instead of the strict optimisation that EUT demands. In this model, people did not calculate the expected utility of every option, but rather used heuristics to make decisions. A variety of research programs developed from here studying heuristics. 

The Simple Heuristics (SH) program founded by Gerd Gigerenzer, researches heuristics that produce good enough choices in contexts where they properly interlock with the environmental affordances and structures. Its premise is that people rely on simple heuristics in conditions of risk and uncertainty \cite{gigerenzer2008heuristics}. Gigerenzer argues that heuristics are as efficient, if not more efficient than utility optimising \cite{gigerenzer2011heuristic} but can fail when applied in the wrong environment, or when faced with information that is profoundly confusing \cite{gigerenzer2007helping}. Good performance is the product of matching the mind’s tools to the current environment (i.e., ecological rationality; \cite{gigerenzer1999simple}). Therefore, effectively determining which heuristics work well in the decision environment leads to advantageous decisions \cite{pachur2013testing}.

The Social Influence research program explores how social norms – beliefs about how others behave and think we should behave – influence behaviour, attitudes and choice \cite{cialdini2004social, sherman2014dual}. Per \citet{cialdini2001science, cialdini2007influence}, when forming judgments people use social heuristics – following a social norm. The SI programme’s assumptions on cognitive architecture is that people are “cognitive misers” who, due to their minds limited cognitive resources, aim to save time and effort when making judgments \cite{fiske1991social}. However, cognitive misers can be motivated to allocate more resources to certain judgments. Per \citet{lieberman2013social}, people’s desire to be liked by members of their social group serves as an important motivational factor.

The Heuristics and Biases (H\&B) programme is rooted in Kahneman and Tversky’s study of people’s decisions in hypothetical financial decisions. They found that people's decisions often violate the axioms of EUT in predictable ways such as the Certainty, Isolation and Reflection effects. Due to these violations, \citet{kahneman1979prospect} developed Prospect Theory (PT) and Cumulative Prospect Theory (CPT; \cite{tversky1992advances}) which posits that people do not make consistently rational decisions across different contexts. CPT describes four factors that influence decision outcomes \cite{kahneman1979prospect}. The factors are: reference dependence - people measure gains and losses relative to a reference point; diminishing sensitivity - people are more sensitive to changes near the reference point; loss aversion – people are more sensitive to losses than to gains of the same magnitude; and probability weighting - people weigh outcomes by decision weights, rather than by their objective probabilities, and thus underweight likely events and overweight unlikely events.

\citet{kahneman2003perspective,kahneman2011thinking} formulated his idea of cognitive architecture, a dual-process account of reasoning. He proposed that people have "two systems" in their mind - System 1 and System 2. System 1 thinking is heuristic. It reacts intuitively and effortlessly, without analysing all available information. System 2 is an analytical and effortful, rationalising process. System 1 thinking is fast, and thus accounts for most behaviour. System 2 can re-evaluate System 1 thinking, thus using System 2 thinking leads to fewer erroneous decision. However, this is difficult, as it requires more cognitive effort. Importantly, some factors and contexts are more likely to trigger System 1 or System 2 thinking than others.

\section{Criticisms of dual-process accounts}
As \citet{newell2015straight} discuss, the dichotomy between System 1 and System 2 is far from uncontroversial in cognitive psychology. Even amongst those proponents of the theory, there is no single unifying account of the features and attributes of System 1 and 2. If we accept the theory, there is further debate about how the systems interact when in conflict \cite{evans2008ar}. Default interventionalists believe there is a serial process where System 1 thinking occurs by default and System 2 is only engaged when a problem is detected. Others propose the two processes occur in parallel leading to conflict. 

More recently it has been suggested that neither approaches are particular satisfactory and some hybrid seems more plausible \cite{deconflict}. For example, in order for the default interventionalist account to hold there needs to be an omnipresent analytic process that can detect a problem. It would be impossible to only activate an analytic process when a problem is detected as the analytic process is necessary in order to detect the problem in the first instance \cite{newell2015straight}. The parallel model on the other hand seems to be cognitively expensive, clashing with other prominent theories such as the Free Energy Principle, which suggest that the mind tries to conserve its energy \cite{friston2010free}. If both intuitive and analytical routes result in the same answer, a parallel model clashes with principles of cognitive economy. The hybrid model thus proposes a ‘hybrid two-stage model’ consisting of a ‘shallow analytic monitoring process’ and an ‘optional deeper processing stage’ \cite{deconflict}. The shallow analytic processing is always engaged while the deeper processing is engaged when there is a conflict between the shallow processing and the intuitive response. Further, the fact that certain experimental tasks appear to lead to a longer response time than others could come from the fact that certain tasks simply require more processing rather than different processing \cite{newell2015straight}. In other words, it might be a matter of different quantities of processing rather than different qualities of processing \cite{evans2008ar}. 

Given that it is difficult to interpret decision-making under uncertainty as resulting from System 1 or System 2, and there are still many uncertainties around how the two systems interact, many have argued against the utility of the theory at all \cite{keren2013tale,osman2013case}. A theory founded on two systems may be impeding on new theoretical progress. 

The purpose of describing the debate surrounding the validity of the two systems is to warn those outside psychology that, despite the intuitive attraction of the theory, the reality is messier. Algorithmic decision makers with System 1 and System 2 characteristics do not gain any additional scientific validity from psychology and the distinction is only as useful as the resulting advantages such a design might lend the decision maker. Fortunately, the computer scientist has rather more control and insight over the decision making process of an algorithm than psychologists enjoy. Algorithmic decision makers really can be designed that operate with two cognitive systems (with the caveat that the characteristics of the two systems are not universally agreed in psychology). 

\section{Intent and System 1 Thinking}

Intentionality does not have a theoretical role in the dual-systems framework \cite{schlosser2019dual}. As regards the particular research question of this article, ‘does model-free thinking admit intent?’, a definitive answer isn’t going to be found from empirical psychological research. Simon (1992) makes the point that intuitive action (System 1 thinking) can be viewed as action motivated by recognition. If a situation has been encountered before it can be recognised, and the behaviour repeated. This is exactly how the policy function that we described earlier works. As \citet{newell2015straight} put it, the state of the world gives the agent a cue to information stored in memory which then provides an answer how to behave. The process can be quick because the steps taken to make that association need not be retraced (sometimes not even voluntarily as we will discuss later). The way that information is stored to produce this ‘situated action’ is through experience which is precisely the way RL agents learn. The experience could have been learned through model based thinking or simple trial and error; both model-free and model-based RL agents produce a policy function.

Although intentionality is not a theoretical concept in the dual-systems account, some have identified intentional actions with controlled, System 2 thinking. Further, researchers claim that many, if not most, of our behaviour stems from System 1 thinking \cite{kahneman2011thinking, gigerenzer2002bounded}. This raises the question of whether intuitive, unconscious, System 1 thinking clashes with the assumption that most of our everyday behaviour can qualify as intentional. \citet{schlosser2019dual} argues that there isn’t a clash and that intention does not exclusively stems from System 2 thinking. Schlosser points out that philosophers of action have already argued for the intentionality of habitual actions, which can be viewed as stemming from System 1 thinking. As put by \citet{davidson2001essays} “we cannot suppose that whenever an agent acts intentionally he goes through a process of deliberation or reasoning.” Intentions can thus be defined in terms of their functional role, and thus it is not necessary that an agent is aware of the intention. As put by \citet{schlosser2019dual} “It is sufficient, that is, if the action is initiated and guided by an intention that is consciously accessible." Because it is possible for habits to be intentional, this provides the basis for seeing actions stemming from system 1 thinking as intentional. 

It is commonly assumed that intentions are based on desires. It thus becomes tempting to view desires as coming from System 1 thinking and the endorsement of desires from System 2 thinking. \citet{schlosser2019dual} argues that not all desires have stem from System 1 thinking, pointing out that many desires are “reason-responsive” – responsive to judgments of System 2 thinking. An example comes from learning that something is unattainable, sometimes causing desires to vanish. There is empirical evidence to support that desires can be reason-responsive \cite{dill2014addict}. \citet{schlosser2019dual} further argues that the endorsement of desire need not only come from System 2 processing. An example comes from “automatic goal pursuit”. By activating certain stimuli, contexts can nudge agents to goal-pursue in a way that is sensitive to the agents’ perceptions of how a goal should be pursued given the circumstances. Thus, automatic goal pursuit – a System 1 process – can be an example of intentions activating automatically. 
Finally, \citet{schlosser2019dual} argues that an agent has to have either had a conscious intention that initiates an action or must have consciously formed a relevant intention at some point in the past. Both System 1 and System 2 thinking can be experienced consciously. Thus, according to this account "intentional action does depend on conscious intention, but the performance of particular intentional actions does not."

\section{The role of intent in defining acceptable, legal, behaviour}
Intent plays an important, multifaceted role in criminal law \cite{simester_five_2021}. Most commonly we think of its presence or absence as determining the culpability of harms caused by an actor. Causing the death of someone intentionally is labelled murder, whilst causing the death of someone by accident, is not (though it may well have some sanction attached to it depending on various factor such as foreseeability). For any given harm, criminal law applies a sliding scale of culpability depending on the strength of the intent of the actor. Intentional harm is worse than harm committed with knowledge, recklessness, or negligence respectively and punished more. This relationship comes under question when we consider the prospect of harms committed by artificial beings, not least because punishing an algorithm poses problems \cite{abbott_punishing_2019}. A popular response amongst engineers is to concentrate on not causing any harm in the first place, AKA `Safe AI'. Another response amongst lawyers is to focus on negligently caused harms, whose definition does not depend on the mental state of the actor but rather a notional `reasonable actor' \cite{abbott_reasonable_2020}.

Neither response is completely satisfactory because intent in criminal law can also play a role in defining the wrongness of certain behaviour. In other words, the wrongness of the act is dictated by the intent of the actor. For example, competition law often prohibits behaviour which is intended to drive other companies out of business. Refusing to identify intent in an algorithm means that certain laws, which were put there for a reason, are unenforceable. The largest category of these offences is those which involve deception of which various definitions exist depending on the type of law being considered and where in the world it applies. Deception most often appears in the various offences concerning fraud. According to \citet{kneer_can_2021} the general folk account of deception includes the following three elements: 1) A representation of a false fact by the agent, 2) knowledge that the representation was false and 3) intent on behalf of the agent that the fact should be believed by the addressee of the representation. As Kneer observes the first two requirements are generally acceptable but the third is somewhat problematic. Nevertheless, as his study shows laypeople seem comfortable labelling deceptive behaviour to algorithms. This is in line with our own experiments (forthcoming!) where we found only small differences in the way people ascribe intent to algorithmic actors.

It would be tempting to think that the role of intent in defining acceptable behaviour is an artefact from when the law only needed to think about humans but legislation continues to refer to the purpose or intent of algorithms. Article 5 of the draft EU AI act prohibits systems that deploy subliminal techniques in order to materially distort a person's behaviour in a manner that causes them harm \cite{franklin_missing_2022}. The recent UK AI regulation policy paper refers both to the purpose and intent of AI systems without elaborating how they might be determined.

Courts do not tend to entertain modish thoughts of human behaviour, nevertheless some legal doctrine considering certain defences might conceivably be recalled when considering the dual-process account. Defences can be divided between those that justify some otherwise criminal activity and excuses which reduce its culpability. Automatism in law refers to behaviour which the accused had no awareness of. This defence has been successfully used to absolve people who have done something in their sleep which would otherwise have been deemed a crime. If it could be argued that model-free RL agents display automatism, then perhaps this means that they cannot act intentionally. Automatism is peculiar because it seems to fall out of the dual-process account altogether. Provocation is an excuse which can reduce the culpability of an action in the situation where an individual has momentarily and reasonably lost control. This most certainly seems to fit the definition of system 1 thinking. It is instructional therefore that provocation can only be used in limited circumstances (typically murder) and modifies not absolves the actor of wrongdoing.

\section{A (legal) definition of intent}
Given the importance that the law places on the cognitive state of the actor at the point of action, how does it choose to define ‘intent’? Unfortunately, the concept has often been left undefined, in common law countries at least, relying on the intuition that a strong folk-concept of intent is shared amongst jurors. This at least has been shown to be the case in a long line of research date back to \cite{malle_folk_1997} albeit with certain biases, most famously the Knobe effect \cite{feltz_knobe_2007}. 
Consider the definition of doing $\phi$ with intention of bringing about outcome X provided by \cite{duff_intention_1990} a) the agent wants or desires outcome X, b) they believe that $\phi$ might bring about X c) they act because of that belief. There is nothing in this account which explicitly requires planning on the part of the agent merely a belief that X is more likely as a result of doing $\phi$. RL agents have a reward function, so it is plausible that if they are directly rewarded for X over and above other outcomes, then they could be said to desire X. If during training, the agent were to find that choosing $\phi$ is more likely to bring about X than $\phi'$, then the probability of choosing $\phi$ would be increased at the expense of $\phi'$. In deployment, it could be argued that since the agent acts according to their policy function and experience has formed their policy to choose $\phi$ which increases the chance of X, both b) and c) are satisfied.
Perhaps we are stretching belief too far. Since a model-free agent has no model they do not know if X is possible by acting in the way that they do when they are in state X’. An alternative formulation that Duff posits is through a test of failure. “An agent intends to bring about those effects whose non-occurrence would (in her eyes at the time of action) render her action a failure.”

\section{Model-free does not imply intent-free}
When a model-free (MF) RL agent acts, and causes a change in the world through their action, they cannot predict what that change might be. On immediate inspection this would seem to pose a problem in the ascription of intent. From a pre-emptive point of view, a MF RL agent cannot tell you alone what it is going to do in the future. This poses obvious safety concerns. Even worse, it seems difficult in a general case to stop such an agent from engaging in intentional behaviour. The stricture "do not lie" for example seems difficult to communicate to a MF RL agent and hard for them to obey. 
Whilst the control problem is somewhat open, we believe the identification of intent in a general RL-agent might not be hinge on whether it has a model of the world. A folk concept of intent has been well studied in experimental psychology \cite{quillien_simple_2021}. This is also relevant for the law because in common law jurisdictions at least where juries comprised of laypeople are used, the legal sense of the word is somewhat tied to the folk sense. Quillien and German test the following definition with good results: An agent intended X through actions $a$ if their attitude towards $X$ caused $X$. Inspired from a legal perspective, \cite{ashton_definitions_2022} suggests an agent intends $X$ through actions $a$ if it is their goal, they had a choice to not $a$, and $a$ foreseeably caused $X$. Both definitions refer to the actor's aims or desires in some sense. We argue these exist in RL agents because they necessarily have a reward function. In many cases, RL agents have a value function which describes the expected reward from being in any state and following their policy thereafter. 
Even with knowledge about a RL agent’s reward function, the task of identifying intended outcomes of an RL agent’s actions is not completely straightforward. Whilst an RL agent could be directly rewarded for causing some harmful outcome by a sociopathic creator, it seems more likely that it would cause harm on the path to completing some legitimate task. We need to take care to specify which of these outcomes can be legitimately labelled side-effects and which can be labelled as intended \cite{ashton_defining_2022}. By labelling certain outcomes as intended (the ends), consistency requires us to label us necessary caused intermediate outcomes (the means) as intended \cite{bratman_intention_2009}. 
By this view, a recommender system designed to maximise user engagement might intend to alter the preferences of its users if that is necessary to achieve its realised engagement \cite{ashton_problem_2022}. This sort of causal analysis requires some external effort to identify intent which the simple MF RL agent cannot conduct, but it seems unsatisfactory to say this implies intent is absent.
The position that model-free RL agents can intend outcomes but the observation that they have no idea about the consequences of their actions does seem contradictory. One possible way to navigate it is to consider RL agents with value functions and use a test for aim based on disappointment. 
We are not the first to make the observation that intent might not require planning in the classic sense. \citet{wallis_intention_2004}, building on the work of  discusses how a non-symbolic Belief, Desire, Intent architecture can be built on the subsumption design of \cite{brooks_intelligence_1991}. This describes a system of intelligence where behaviour is a direct function of an agent’s environment. This is exactly how a policy function directs an RL agent. Wallis makes the point that under this architecture intelligence becomes a function of the agent and the environment that it is in. 

\section{Explaining actions and controlling for intent requires a model}
Whilst the model-free RL agent might be capable of intentional action, the problem remains that without a model of the world they will find it hard to explain why they have chosen their actions and what the implications of those actions will be. Strategies learned through model-free RL are strategies learned through situated action as \cite{simon_what_1992} would put it. Once the policy function is divorced from the environment it is learned in, it becomes extremely difficult for anyone to explain. Simon illustrates this point when discussing someone explaining their solution to the Tower of Hanoi puzzle. Absent a model or an actual physical puzzle in front of her, she cannot proceed to explain the solution. Likewise, her friends, require either their own mental model or puzzle themselves to form an explanation of her behaviour. From the legal definition of intent, because it is a mental state held by the actor, it is clear we should judge intent using the model or means that the actor used to form their strategy. In the case of a model-free RL, this means using the training data (or generator) that produced their policy function. All of this points to the possible conclusion that model-free RL agents can only be called intentional when considered in conjunction with their training environment. 

It is likely that determining the intent behind an algorithm’s actions will become an important task for AI Safety. The implication of this article is that model-free RL agents are not capable of explaining their own actions. Explainability technology will require a model of the world to parse behaviour. For it to be truly accurate, it requires access to the same training environment that the agent developed their policy function. One reason why juries are trusted to judge the intent of others is that it is assumed humans share a common understanding of the world. This assumption is broken when humans or algorithms are asked to judge the intent behind the actions of algorithmic agents. A parallel can be drawn with the dual process account; some argue that humans’ casual explanatory thinking ability came about to attempt to explain to ourselves what our intuitive System 1 is doing. 

A problem with an model-free RL agent being unable to explain its own behaviour is that it cannot control certain elements of its behaviour. We cannot ask for it to not intend to do something (excuse the double negative). Controlling the behaviour or outcomes of an RL agent is a topic taken up by Safe RL \cite{garcia_comprehensive_2015}. Whilst originally safety was narrowly construed as a restriction on reaching certain states, more recently this has enlarged to recognise that we will often want to restrict behaviours as well as outcomes \cite{balakrishnan_incorporating_2019,hasanbeig_logically-constrained_2019}.

An interesting technique developed by the field is that of the shield \cite{alshiekh_safe_2017,jansen_safe_2019}. This is a structure that sits between the RL-agent and the environment and using information from both, ‘shields’ certain actions from being chosen by the agent at certain states of the world in order to enforce some exogenous restriction on behaviour. The shield has the effect of forcing the RL-agent to learn safe policies during training (because it only ever explores safe policies). Because it can only guarantee safe behaviour in situations experience during training, the shield can also be maintained in deployment, to make sure the agent’s behaviour remains safe when unfamiliar situations are encountered. Techniques vary, but the shield will often have access to a model of the world (or construct an approximation through a MDP (Markov Decision Process)). It requires this to test the behavioural restrictions it is tasked with enforcing. These restrictions are intertemporal so it will use the model to project forward and back to test whether the current policy is or can obey them. Here one can draw a parallel to the System 1 and 2 model of thinking. Whilst the shield is often conceived as something separate to the RL-agent, one could choose to interpret it as a different cognitive system within the agent. 

\section{Conclusion}
In this article we argue that because understanding the intent of an actor is unavoidably useful when defining acceptable behaviour, its study in reinforcement learning agents is important wherever they might be engaged in regulated activity.

Model free (MF) RL agents learn their policies through vast amounts of experience which is then distilled into a policy function. Through their policy function, MF RL agents react to states with actions with no introspection. Nevertheless the policy that they have learned was motivated by an objective function or an aim. Model-free RL agents are purely System 1 thinkers and instinctively this would seem to mean that they cannot engage in intentional behaviours. We describe the origins of the dual-process account and discuss some of its current controversies. We also look at how intent is conceptualised in criminal law and folk psychology. We conclude that psychological and legal thinking does not exclude intent in a model-free agent as long as we consider intent in the wider context of the agent’s policy and the environment in which it was learned.

We conclude by arguing that a model of the world is vitally important in the explanation of behaviour and therefore the identification of intent. This application is important in AI Safety where we want to control the intent of an agent.

\bibliography{aaai22} 

\end{document}